\documentclass{article}

    \PassOptionsToPackage{numbers, compress}{natbib}

\usepackage[preprint]{neurips_2019}

\usepackage[utf8]{inputenc} 
\usepackage[T1]{fontenc}    
\usepackage{hyperref}       
\usepackage{url}            
\usepackage{booktabs}       
\usepackage{amsfonts}       
\usepackage{nicefrac}       
\usepackage{microtype}      
\usepackage{amsmath}
\usepackage{amssymb}
\usepackage{xcolor}
\usepackage{algorithm}
\usepackage{algorithmic}
\usepackage{booktabs} 
\usepackage{subcaption}
\usepackage{listings}
\usepackage{graphicx}
\usepackage{wrapfig,lipsum,booktabs}
\usepackage{color}
\usepackage[serbian,english]{babel}

\newtheorem{definition}{Definition}

\title{Disentangled State Space Representations}

\author{%
  {\DJ}or{\dj}e Miladinovi{\'c} $^{1}$ \quad Muhammad Waleed Gondal $^{2}$ \quad Bernhard Sch{\"o}lkopf $^{2}$ \\ 
  \bf{Joachim M. Buhmann $^{1}$ \quad Stefan Bauer $^{2}$} \\
  $^{1}$ Department of Computer Science, ETH Zurich \\
  $^{2}$ Max-Planck Institute for Intelligent Systems \\
  \texttt{djordjem@ethz.ch}
}

\begin{document}

\maketitle

\begin{abstract}
Sequential data often originates from diverse \emph{domains} across which statistical regularities and domain specifics exist.
To specifically learn cross-domain sequence representations, we introduce \emph{disentangled state space models (DSSM)} -- a class of SSM in which domain-invariant state dynamics is explicitly \emph{disentangled} from domain-specific information governing that dynamics. 
We analyze how such separation can improve knowledge transfer to new domains, and enable robust prediction, sequence manipulation and domain characterization.
We furthermore propose an unsupervised VAE-based training procedure to implement DSSM in form of Bayesian filters.
In our experiments, we applied VAE-DSSM framework to achieve competitive performance in online ODE system identification and regression across experimental settings, and controlled generation and prediction of bouncing ball video sequences across varying gravitational influences.
\end{abstract}

\section{Introduction}

\begin{figure}[!t] 
	\centering	
	\includegraphics[width=\textwidth]{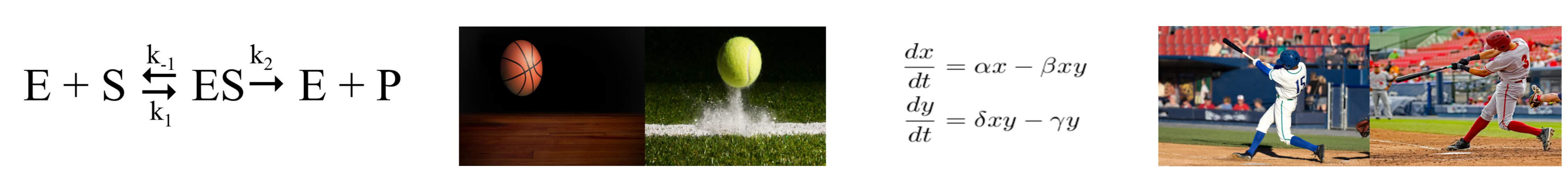}
	\caption{\textbf{Sequence dynamics across domains.} From left to right: \emph{(i)} Michaelis-Menten model for enzyme kinetics, governed by reaction rate constants $\vec{k}$; \emph{(ii)} bouncing ball kinematics, determined by ball weight and playground characteristics; \emph{(iii)} ODE dynamics, governed by model parameters; \emph{(iv)} bat swinging motion, influenced by the person performing it.
	Generic dynamics is defined by a domain-invariant mechanism, but differences appear due to domain-specific factors of variation.}
	\label{fig-confounded-dynamics}
	\vspace{-1em}
\end{figure}

Learning sequence dynamics and representations is a major challenge in many areas of machine learning and beyond.
Albeit successful in practice, current sequence models (e.g. recurrent neural networks) do not \emph{per se} acknowledge the existence of \emph{domains} across which sequences are not IID\footnote{IID -- independent and identically distributed.}.
Namely, in many practical scenarios (see Figure \ref{fig-confounded-dynamics}), sequence dynamics $X_i|\vec{X}_{<i}$ differs across domains $D$ which we denote as  $P^{D_k}(X_i|\vec{X}_{<i}) \neq P^{D_m}(X_i|\vec{X}_{<i})$ for $k \neq m$.
In this work, we propose to independently model and learn domain recognition posterior $P(D|\vec{X}_{<i})$ to capture domain-relevant information, and domain-invariant conditional $P(X_i|\vec{X}_{<i},D)$ to capture generic regularities.
From the predictive modeling perspective, we analyze how this can facilitate knowledge transfer to previously unseen sequences.
From the representation learning point of view, we show that the proposed model can be used to characterize sequences, manipulate their dynamics, and even isolate independent sources of variation which appear across domains.

How to learn such flexible, domain-aware sequence representations?
We build on recent advances in learning of non-parametric~\emph{state space models (SSM)}~\citep{karl2016deep}.
On one hand, these model-free SSM offer more flexibility (e.g. can model video sequences) than the traditional engineering SSM which are typically found in form of (usually linear) carrefully crafted Kalman filters~\citep{gelb1974applied}.
On the other hand, due to their non-autoregressive architecture they make an attractive alternative to recurrent neural networks in data rich settings.
In particular, we present a novel class of SSM crafted for multi-domain sequential data.
Several recent works have already recognized the benefits of introducing additional structure into SSM: the requirement of separating confounders from actions, observations and rewards \citep{lu2018deconfounding} or content from dynamics \citep{yingzhen2018disentangled,fraccaro2017disentangled}. Complementary to these approaches, we focus on learning structured SSM to \emph{disentangle} sequence dynamics into its generic (domain-invariant) and domain-specific factors.

\subsubsection*{Our key contributions:}

\begin{figure*}[htbp]
    \begin{center}
    \centerline{\includegraphics[width=1\linewidth]{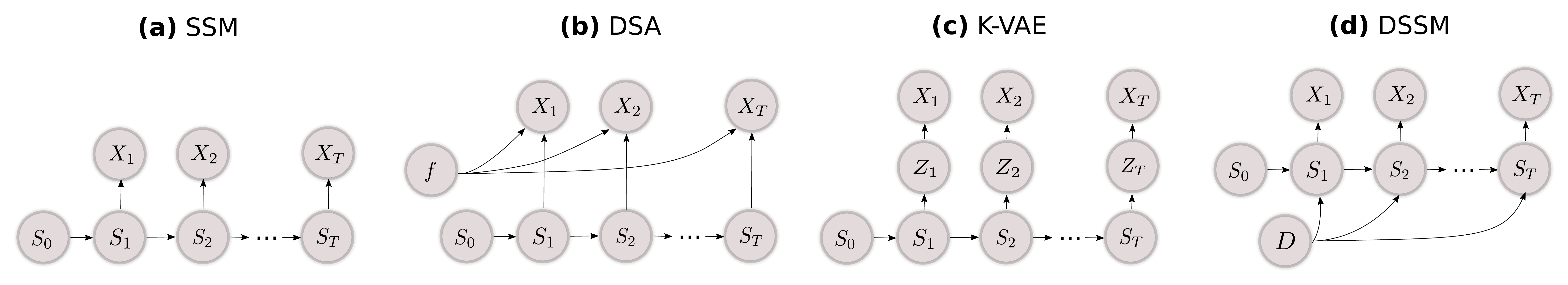}}
    \caption{\textbf{DSSM and related architectures.} \textbf{(a)} Traditional SSM architecture was used e.g. by~\cite{karl2016deep}. \textbf{(b)} Disentangled sequential autoencoder (DSA)~\citep{yingzhen2018disentangled} decouples time-invariant \emph{content} from the time-varying features. \textbf{(c)} Kalman-VAE~\citep{fraccaro2017disentangled} separates object (content) representation from its dynamics (we did not depict control input here). \textbf{(d)} DSSM introduce domains $D$ to model domain-specific effects on sequence \emph{dynamics}.
    }
    \label{fig-graphical-model}
    \end{center}
    \vspace{-1.5em}
\end{figure*}

\begin{itemize}
    \item \textbf{Cross-domain sequence dynamics learning with DSSM.} We formulate and discuss the problem of learning sequence models from heterogeneous domains. We then introduce a class of non-parametric SSM tailored to this problem -- \emph{disentangled state space models (DSSM)} (depicted in  Figure~\ref{fig-graphical-model}d) form a joint domain model while explicitly decoupling what is generic in sequence dynamics from what is domain-specific.
    \item \textbf{Variational Bayesian filtering for DSSM.} We extend on recent advances in amortized variational inference to design an unsupervised training procedure and implement DSSM in form of Bayesian filters.
    Similarly to~\cite{karl2016deep}, well-established reparameterization trick is applied such that the gradient propagates through time.
    \item \textbf{ODE learning.}  As our first application, we learn an ODE system from varying experimental configurations (domains).
    In contrast to current state-of-the-art, with \emph{no prior knowledge} of the actual ODE form, our method remarkably \emph{recovers true parameters} -- from the previously unseen test sequence. 
    This in turn facilitates learning of a robust Bayesian filter that \emph{extrapolates well} on a long horizon and from very noisy observations.
    As opposed to current solutions which require hours, this is done \emph{online}.
    \item \textbf{Video prediction and manipulation.} We analyze video sequences of a bouncing ball, influenced by varying gravity (domain).
    We outperform state-of-the-art in predictions, and also \emph{do interventions} by "swapping domains" i.e. we enforce a specific dynamic behaviour by using a domain from another sequence which exhibits the desired behaviour.
    Example videos are available at: \url{https://sites.google.com/view/dssm}.
\end{itemize}

\section{Related Work}
\label{sec:rw}

Broadly speaking, our efforts go in line with a longstanding promise of artificial intelligence to design agents that can extrapolate experience across scenarios. 
Research community has embodied some of these efforts into concepts such as domain adaptation and transfer learning~\citep{pan2010survey}, catastrophic forgetting~\citep{french1999catastrophic}, and learning of disentangled representations~\citep{bengio2013representation}.
In line with recent findings \citep{locatello2018challenging, locatello2019disentangling} which emphasize the vital role of inductive biases for obtaining well-disentangled representations, we leverage on temporal coherence and domain stationarity to regularize our learning procedure.
In some sense, our work can be seen as an attempt to address the topics of transfer learning and disentanglement in dynamic data settings.

VAE frameworks have already been extended to sequence modeling~\citep{marino2018general}, and applied to speech~\citep{bayer2014learning,chung2015recurrent,fraccaro2016sequential,goyal2017z}, videos~\citep{yingzhen2018disentangled} and text~\citep{bowman2015generating}.
However, these (mainly) recurrent neural network-based approaches are autoregressive and hence not always suitable e.g. for planning and control from raw pixel space~\citep{hafner2018learning}. 
To "image the world"~\citep{ha2018world} from the latent space directly and circumvent the autoregressive feedback, alternative methods learn SSM instead~\citep{karl2016deep,fraccaro2017disentangled,krishnan2017structured}.
In DVBF~\citep{karl2016deep}, SSM is trained using VAE-based learning
procedure which allows gradient to propagate through time during training. 
K-VAE~\citep{fraccaro2017disentangled}  is a two-layered model which decomposes object's representation from its dynamics.
DKF~\citep{krishnan2015deep}, and very closely related DMM~\citep{krishnan2017structured}, admit SSM structure but the state inference is conditioned on both past and future observations, so the structure of a filter is not preserved.
This is problematic as noted by~\cite{karl2016deep}. Similar issues can be found in DSA~\citep{yingzhen2018disentangled}.

Closely related to ours are approaches which consider structured and disentangled representation of videos, separating the pose from the content~\citep{denton2017unsupervised,tulyakov2017mocogan,villegas2017decomposing,yingzhen2018disentangled} or the object appearance from its dynamics~\citep{fraccaro2017disentangled}.
Proposed models were shown to improve the prediction~\citep{villegas2017decomposing,denton2017unsupervised} and enable controlled generation with "feature swapping"~\citep{tulyakov2017mocogan,yingzhen2018disentangled}.
Similar \emph{content-based disentanglement} was proposed in speech analysis for separating sequence- and segment-level attributes~\citep{hsu2017unsupervised}.
In contrast to \emph{content-based} methods which focus exclusively on the observation model, our work is more general and can adapt to cross-domain variations in sequence dynamics, effectively performing \emph{dynamics-based} disentanglement.
In this regard, our approach can be seen as complementary to existing (see also Figure~\ref{fig-graphical-model}).
For example, while DSA can represent and manipulate the shape or color of a bouncing ball, our method can manipulate its trajectory in addition (see our video experiments).

\section{Cross-domain Sequence Dynamics Analysis with DSSM}
\label{sec:theory}

We consider a setting in which we are given $N$ sequences $\vec{X}^{k=1..N}$ describing equivalent time-evolving phenomena, but each originating from a different domain $D_k$.
We assume that sequence dynamics may vary across domains: $P^{D_k}(X_i|\vec{X}_{<i}) \neq P^{D_m}(X_i|\vec{X}_{<i})$ for $k \neq m$\footnote{Alternatively, one can equivalently formulate non-IID setting as: $P^{D_k}(\vec{X}) \neq P^{D_m}(\vec{X})$ for $k \neq m$.}.
We further assume there exist some shared knowledge i.e. statistical relations in sequence dynamics valid across all domains.
Our task is to learn a robust cross-domain sequence model $P(X_i|\vec{X}_{<i})$ which can extrapolate knowledge on new, possibly unseen domains.
We are also interested in exploring explicit differences between domains, to characterize them, interpolate between them, observe their influence on generic sequence dynamics, and explain independent sources of their variability.

Our main idea is based on explicit construction of the latent domain space $\mathcal{D}$ and corresponding sequence/domain embeddings $D \in \mathcal{D}$.
We would like to learn compact $\mathcal{D}$, such that similar sequences are stored close to each other.
This would imply that when a new sequence appears, we can implicitly transfer knowledge from the sequences which have similar embeddings.
$D$ should ideally contain only relevant information for emulating domain-specific sequence dynamics discarding all extraneous details.
During training we simultaneously learn two mechanisms, the first one used to embed sequences into $\mathcal{D}$, and the second one to capture general regularities in sequence dynamics:
\begin{align}
    \vec{X}_{<i} \rightarrow& D \; \text{\emph{(domain recognition mechanism)}} \label{eq:dr}\\
    \vec{X}_{<i},D_k \rightarrow& {X}_i \; \text{\emph{(domain-invariant transition mechanism)}} \label{eq:di}
\end{align}
To ensure $D$ captures stationary information, it is kept constant within a sequence, influencing transition at each time step.
In order to enable modeling of sequences of arbitrary complexity in structure and dynamics, we first define a special class of flexible non-paramateric SSM and then describe corresponding amortized variational inference-based training procedure for parallel learning of two mechanisms given in Eq (\ref{eq:dr}) and Eq (\ref{eq:di}).
\begin{definition}[DSSM]\label{def-dssm} Let $S_i$ and $X_i$ be probabilistic representations of the latent state and observation at time step $i$.
Let $\omega_i$ and $\beta_i$ be probabilistic random variables describing observation and latent process noise.
Furthermore, we define $D$ as a latent variable describing the domain of modeled phenomena.
A disentangled state space model is described as a stochastic process in which these variables are related as follows:
\begin{align}
X_{i} &= g(S_i, \omega_i) \label{eq-observation-model} \\
S_{i+1} &= f(S_i, D, \beta_i) \label{eq-transition-model}
\end{align}
where $g$ and $f$ are observation and transition (deterministic) functions of arbitrary form. 
\end{definition}

\section{Variational Bayesian filtering for DSSM} \label{sec-bayes-filter}

\begin{figure*}[!b]
    \begin{center}
    \centerline{\includegraphics[width=\linewidth]{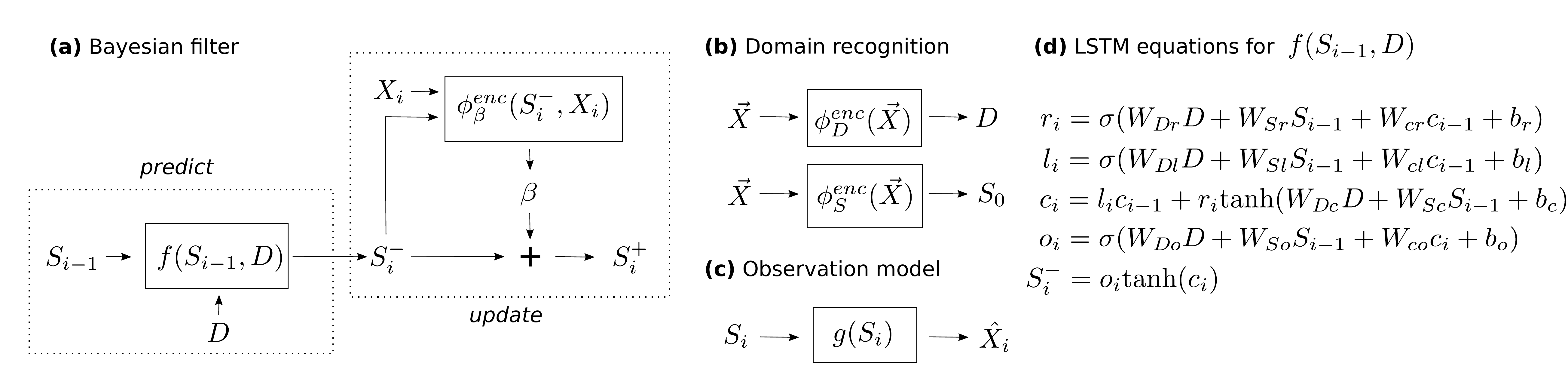}}
    \caption{\textbf{Variational Bayesian filtering framework.} 
    \textbf{(a)} predict-update equations of proposed Bayesian filer. \emph{Predict step} generates the \emph{a priori} state estimate $S_i^-$. \emph{Update step} corrects the estimate ($S_i^+$) using observation $X_i$ through the residual vector $\beta_i$; \textbf{(b)} in domain recognition phase the domain $D$ is inferred, and the initial state $S_0$; \textbf{(c)} function $g$ maps states into observations; \textbf{(d)} an example implementation of the transition function $f$ from Eq (\ref{eq-det-transition}) using LSTM equations~\citep{graves2013generating}..
    }
    \label{fig-bayes-filter}
    \vspace{-2em}
    \end{center}
\end{figure*}

In this work, we focus on deterministic systems assuming that the latent process noise $\beta$ and observation noise $\omega$ are both uncorrelated in time.
Hence we consider the following version of Def~\ref{def-dssm}:
\begin{align}
    X_{i} &= g(S_i) + \omega_i, \quad \omega_i \sim N(0,\Sigma_{\omega}) \label{eq-det-observation}\\
    S_{i+1} &= f(S_i, D) + \beta_i, \quad \beta_i \sim N(0,\Sigma_{\beta}) \label{eq-det-transition}
\end{align}
$X_i\in \mathbb{R}^O$ represents the $O$-dimensional observation in time step $i$ and $S_i\in \mathbb{R}^L$ is the $L$-dimensional latent state.
$\Sigma_{\beta}\in \mathbb{R}^{L\times L}$ and $\Sigma_{\omega}\in \mathbb{R}^{O\times O}$ are the process and observation noise covariances which we will for simplicity assume are isotropic Gaussian.
Our goal will be to jointly learn cross-domain generative model which includes the transition function $f$ and the observation function $g$, and also the corresponding recognition networks $\phi_{\beta}^{enc}$, $\phi_{D}^{enc}$ and $\phi_{S}^{enc}$ which infer the process noise residual $\beta_i$ (similarly to ~\cite{karl2016deep}), domain $D$, and initial state $S_0$ respectively.
The overview is given in Figure~\ref{fig-bayes-filter}.
 
\paragraph{Generative model.} Given an observed sequence $\vec{X}$ of length $T$, the joint distribution is:
\begin{align}
&p(\vec{X}, \vec{S}, D, \vec{\beta}) = p_0(S_0) p_0(D) \prod_{i=1}^{T} p(X_i|S_i) p(S_i| S_{i-1}, D, \beta_i) p_0(\beta_i) \label{eq-full-gen}
\end{align}
This follows from Figure~\ref{fig-graphical-model}d and the assumption that the process noise is serially uncorrelated. 
We set the prior probabilities of the initial state $p_0(S_0)$, domain $p_0(D)$ and process noise $p_0(\beta_i) = p_0(\beta)$ to be zero-mean unit-variance Gaussians.
Conditioned on $\beta_i$ and $D$, state transition is deterministic i.e. the probability $p(S_i| S_{i-1}, D, \beta_i)$ is a Dirac function with the peak defined by Eq~(\ref{eq-det-transition}). The emission probability $p(X_i|S_i)$ is defined by Eq~(\ref{eq-det-observation}).

\paragraph{Inference.} 

Joint variational distribution over the unobserved random variables $D$, $\vec{S}$ and $\vec{\beta}$, for a sequence of observation $\vec{X}$ of length T factorizes as:
\begin{align}
q(\vec{S}, D, \vec{\beta}|\vec{X}) &=  q(D|\vec{X})q(S_0|\vec{X}) \prod_{i=1}^{T} q(S_i|\beta_i,S_i^-) q(\beta_i|S_i^-, X_i) q(S_i^-|S_{i-1},D)
\end{align}
Here, the conditionals $S_i^-|S_{i-1},D$ and $S_i|\beta_i,S_i^-$ are deterministic and defined by Eq (\ref{eq-det-transition}). The remaining factors are given as follows:
\begin{align}
q(\beta_i|S_i^-, X_i) &= \mathcal{N}(\mu^{\beta_i},\Sigma^{\beta_i}), \quad [\mu^{\beta_i}, \Sigma^{\beta_i}] = \phi^{enc}_\beta(S_i^-,X_{i}) \\
q(S_0|\vec{X}) &= \mathcal{N}(\mu^s, \Sigma^S), \quad [\mu^S, \Sigma^S] = \phi_S^{enc}(\vec{X}) \\
q(D|\vec{X}) &= \mathcal{N}(\mu^D, \Sigma^D), \quad [\mu^D, \Sigma^D] = \phi_D^{enc}(\vec{X})
\end{align}

\paragraph{Learning.}

To match the posterior distributions of $D$, $S_0$ and $\vec{\beta}$ to the assigned prior probabilities $p_0(D)$, $p_0(S_0)$ and $p_0(\beta)$, we utilize reparametrization trick~\citep{kingma2013auto, rezende2014stochastic}. This enables end-to-end training.
To define the objective function we derive the variational lower bound $\mathcal{L}$ which we consequently attempt to maximize during the training.
We start from the well-known equality~\citep{kingma2013auto}:
\begin{align}
\mathcal{L} = \mathbb{E}_{q(\vec{S}|\vec{X})}[\text{log } p(\vec{X}|\vec{S})] - \text{KL}(q(\vec{S}|\vec{X})||p_0(\vec{S})) \label{eq-lb}
\end{align}
Due to the conditional independence of the observations given the latent states, we can decompose the first term as:
\begin{align}
\mathbb{E}_{q(\vec{S}|\vec{X})}[\text{log } p(\vec{X}|\vec{S})] = \sum_{i=1}^T \mathbb{E}_{q(S_i|\vec{X})}[\text{log }p(X_i|S_i)] \label{eq-ll}
\end{align}
The KL term can be shown to simplify into a sum of the following KL terms:
\begin{align}
\text{KL}(q(\vec{S}|\vec{X})||p_0(\vec{S})) &= \text{KL}(q(D|\vec{X})||p_0(D))  \label{eq-kl} \\
                                            &+ \text{KL}(q(s_0|\vec{X})||p_0(S_0)) \nonumber\\
                                            &+ \sum_{i=1}^T \underset{q(\beta_i,D,S_{i-1}|\vec{X})}{\mathbb{E}}\text{KL}(q(\beta_i)||p_0(\beta)) \nonumber
\end{align}
where we dropped the conditional dependency $\beta_i |S_{i-1},X_i,D$ in $q$ to ease the notation.
Full $\mathcal{L}$ derivation is given in Appendix~\ref{sec-lowerbound}. Algorithm~\ref{alg-training} in Appendix~\ref{a:la} shows the details of the training procedure for one iteration, for a batch of size 1. The extension to larger batches is trivial.

\paragraph{Implementation.}

We model [$f$, $g$, $\phi_S^{enc}$, $\phi_D^{enc}$,  $\phi_\beta^{enc}$] as neural networks, with slightly varied architecture depending on the data. 
In particular, $g$ and $\phi_\beta^{enc}$ may be given as multi layer perceptrons or convolutional/deconvolutional networks (see experiments for details).
Common to all experiments, $\phi_S^{enc}$ and $\phi_D^{enc}$ are bi-directional LSTM ~\citep{graves2013generating} followed by a multilayer perceptron to convert LSTM-based sequence embedding into $S_0$ and $D$. $f$ is modeled as LSTM cell as elaborated in Figure~\ref{fig-bayes-filter}d.

\paragraph{Optimization tricks.}
We observed improved domain recognition inference with an additional heuristic regularization term. 
Since $D$ models the time-invariant, global information of the state dynamics, and by design (Figure~\ref{fig-graphical-model}d) influences state transition in each time step, we penalize the step-wise change of the hidden representations of the LSTM which models $\phi_D^{enc}$. To that end, we define an additional \emph{moment matching regularization} term as:
\begin{align}
MM(\phi_D^{enc}(\vec{x})) = \sum_{i=2}^{T}||h_{i} - h_{i-1}||^2 \label{eq-mm}
\end{align}
where $h_i$ is the hidden state of the $\phi_D^{enc}$ LSTM cell in step $i$. This idea is related to the approaches based on the maximum mean discrepancy~\citep{gretton2012kernel}.
Namely, enforcing equality of consecutive cell states corresponds to matching of their first moments.
Secondly, similarly to~\cite{bowman2015generating,karl2016deep} we used a KL annealing scheme. 
This was helpful for circumventing local minimum and preventing the KL term to converge to zero too early during the training.
The exact details are given in our experiments.

\paragraph{Disentangling independent factors of variation.} 
We leverage on the intrinsic property of VAE to shape compact latent spaces. 
In particular, in the spirit of $\beta$-VAE~\citep{higgins2017beta}, we introduce an additional coefficient which we call $\delta$, to multiply KL divergence domain embedding term -- the first one in Eq (\ref{eq-kl}).
Like in $\beta$-VAE, the idea is to encourage the decomposition of the latent space and isolate underlying generative factors of variation.

\section{Experiment -- Learning ODE Dynamics}
\label{sec:dynamical_system}

In many real-world applications e.g describing gene regulatory interactions \citep{calderhead2009accelerating} or brain connectivity analysis \citep{friston2003dynamic}, regression and parameter inference in ordinary differential equations (ODE) is of crucial importance.
We analyze how DSSM, perhaps somewhat surprisingly, can be used for both tasks.
We train DSSM on a large set of simulated sequences with varying parameter configurations, and then test on a previously unseen sequence.
We then study how trained DSSM disentangles domain-specific information (ODE parameters) from generic regularity (ODE functional form).

\paragraph{Lotka-Volterra ODE system.} Predator-pray model was originally introduced in the theory of autocatalytic chemical reactions \citep{lotka1910}. Mathematically, it is described as follows:
\begin{align}
\dot{x}_1(t) &= \alpha_1 x_1(t) - \alpha_2 x_1(t) x_2(t) \label{eq-lv1}\\
\dot{x}_2(t) &= - \alpha_3 x_2(t) + \alpha_4 x_1(t) x_2(t) \label{eq-lv2}
\end{align}
\paragraph{Data set.} 

We simulated 10'000 sequences for training and validation (5\% for early stopping) by uniformly sampling ODE parameters from $[0.5,4.5]$ range. 
The observations were additionally corrupted with white Gaussian noise with standard deviation of 0.5.
For all sequences, initial states were set to $\mathbf{X}_{1-2}(t=0) = [5,3]$. 50 equidistant points per sequence were "observed".
Following the previous work in ODE inference of parameters and states~\citep{gorbach2017scalable,wenk2018fast}, we generated a benchmark trajectory using parameters $\mathbf{\alpha}_{1-4}$ = [2,1,4,1] (excluded from training).
During test phase, we used 50 first points for system identification (domain recognition), and other 150 to evaluate long-term prediction.
We additionally corrupted test sequence with varying noise magnitudes, with standard deviations of [0.1,0.5,1] to study robustness.
Sequences were generated using Runge-Kutta integrator of order 5 with a time step of 0.01.

\paragraph{Prediction.}

\begin{figure*}[htbp]
        \begin{center}
        \centerline{\includegraphics[width=.95\textwidth]{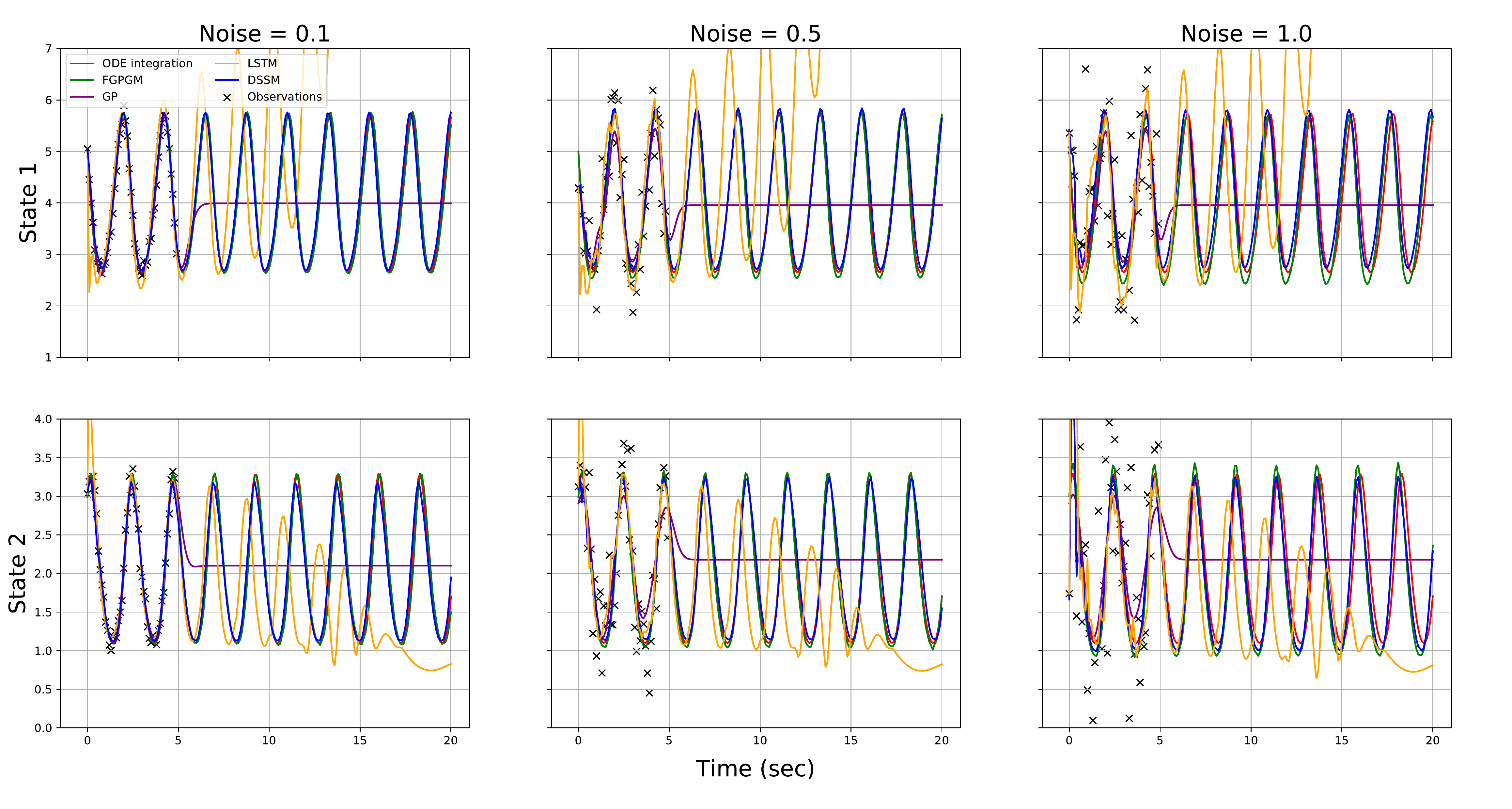}}
        \vspace{-0.5em}
        \caption{\textbf{Prediction on Lotka-Volterra.} Comparison of our method against the baselines for three different noise levels on a standard benchmark trajectory. Shown is DSSM ($\delta=2$). Other baselines were left out to keep the figure uncluttered. 
        }
        \label{fig-noisy-observations}
        \end{center}
        \vspace{-1em}
\end{figure*}

In Figure~\ref{fig-noisy-observations} and Table~\ref{tab-mse} we present the comparison to the baseline methods, for three different noise realisations of increasing magnitude.
We trained \emph{LSTM} as implemented in~\cite{graves2013generating} using \emph{teacher forcing} strategy. 
The offline approaches include classical \emph{Gaussian process (GP)}~\citep{rasmussen2004gaussian} with radial basis function kernel, and its ODE-tailored enhancement, the \emph{FGPGM} method~\citep{wenk2018fast} which regularizes GP estimates using the known functional form of the underlying dynamical system.
As an ablation test, we trained a domain-free version of DSSM (denoted as SSM) simply by fixing $D$ to 0.
DSSM was trained for different values of $\delta$, to demonstrate its positive effects on the model regularization and consequently better transfer learning.
Missing details are found in Appendix~\ref{a:lv}.
Our method shows robustness to very noisy observations, and is clearly able to learn the intrinsic periodical pattern of the dynamical system. 
We observe that DSSM work comparably well in comparison to highly optimized ODE competitor with two crucial advantages: \emph{(A) online estimation}; \emph{(B) model-agnosticity}.
On the other hand, both LSTM and SSM have difficulties learning stable dynamics, and the predictions quickly drift away, while GP does not extrapolate at all.

\begin{figure}[htbp]
    \centering
    \begin{subfigure}[t]{.24\textwidth}
      \includegraphics[width=\linewidth]{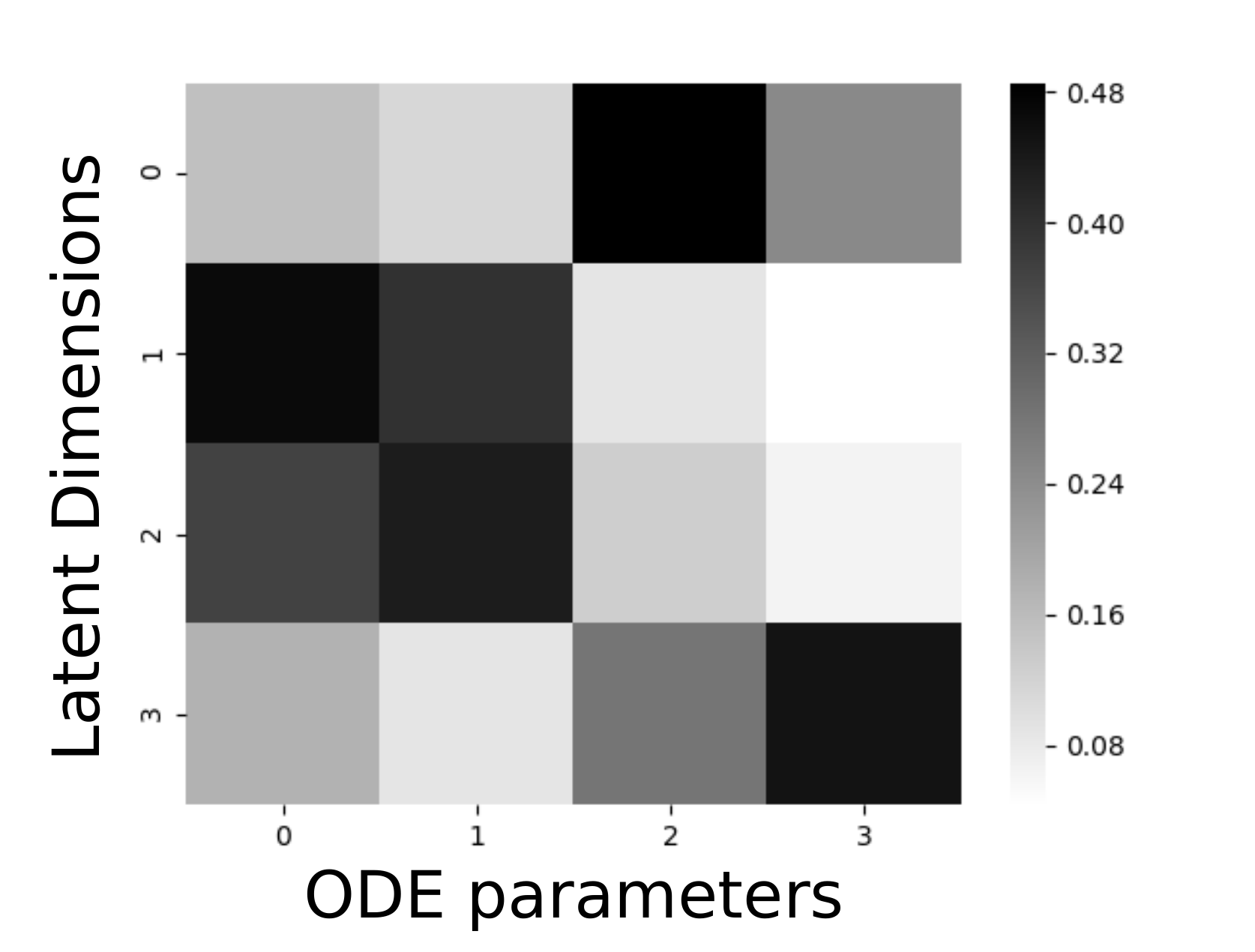}
      \caption{DSSM ($\delta=0$)}
    \end{subfigure}\hfill
    \begin{subfigure}[t]{.24\textwidth}
      \includegraphics[width=\linewidth]{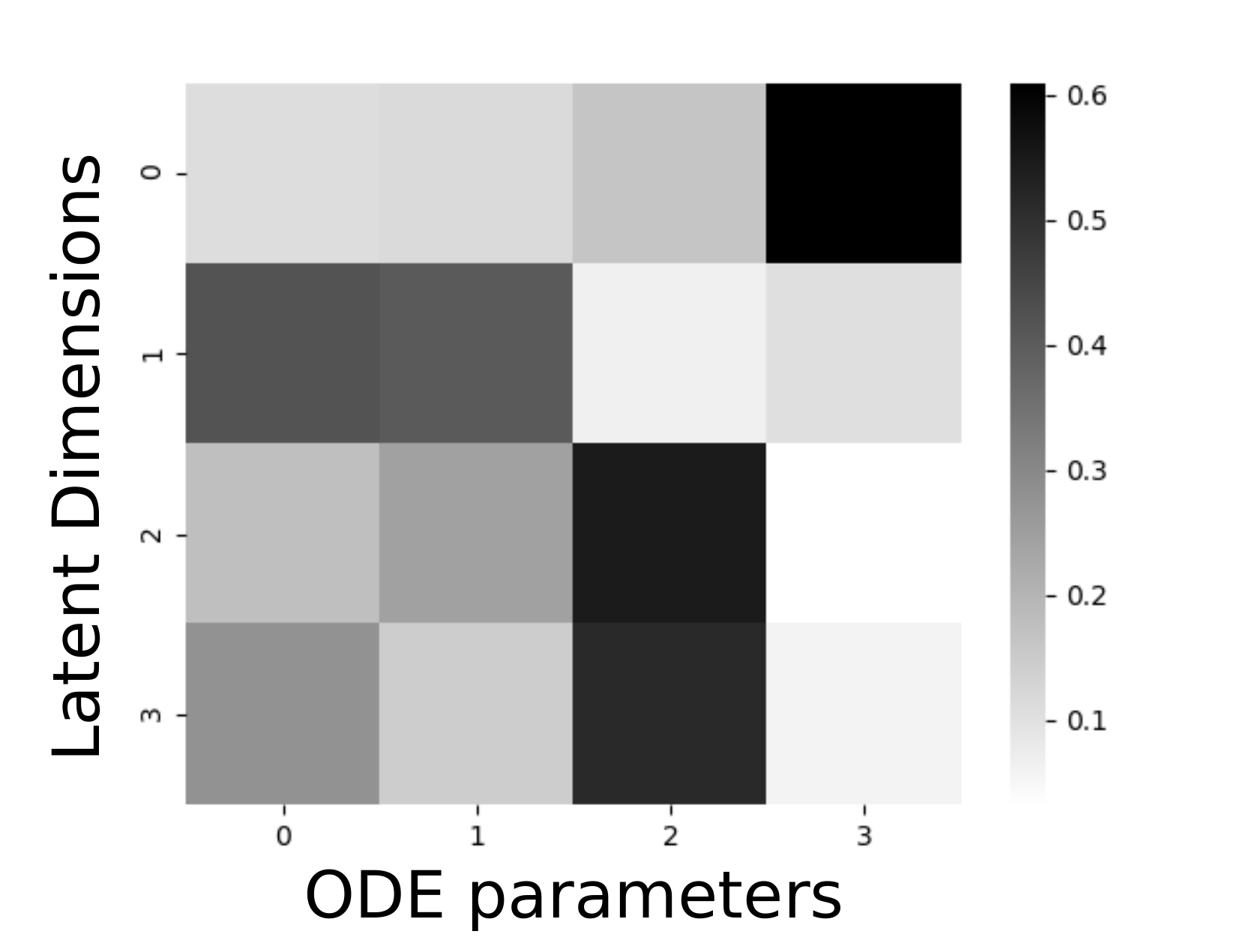}
      \caption{DSSM ($\delta=0.5$)}
    \end{subfigure}\hfill
    \begin{subfigure}[t]{.24\textwidth}
      \includegraphics[width=\linewidth]{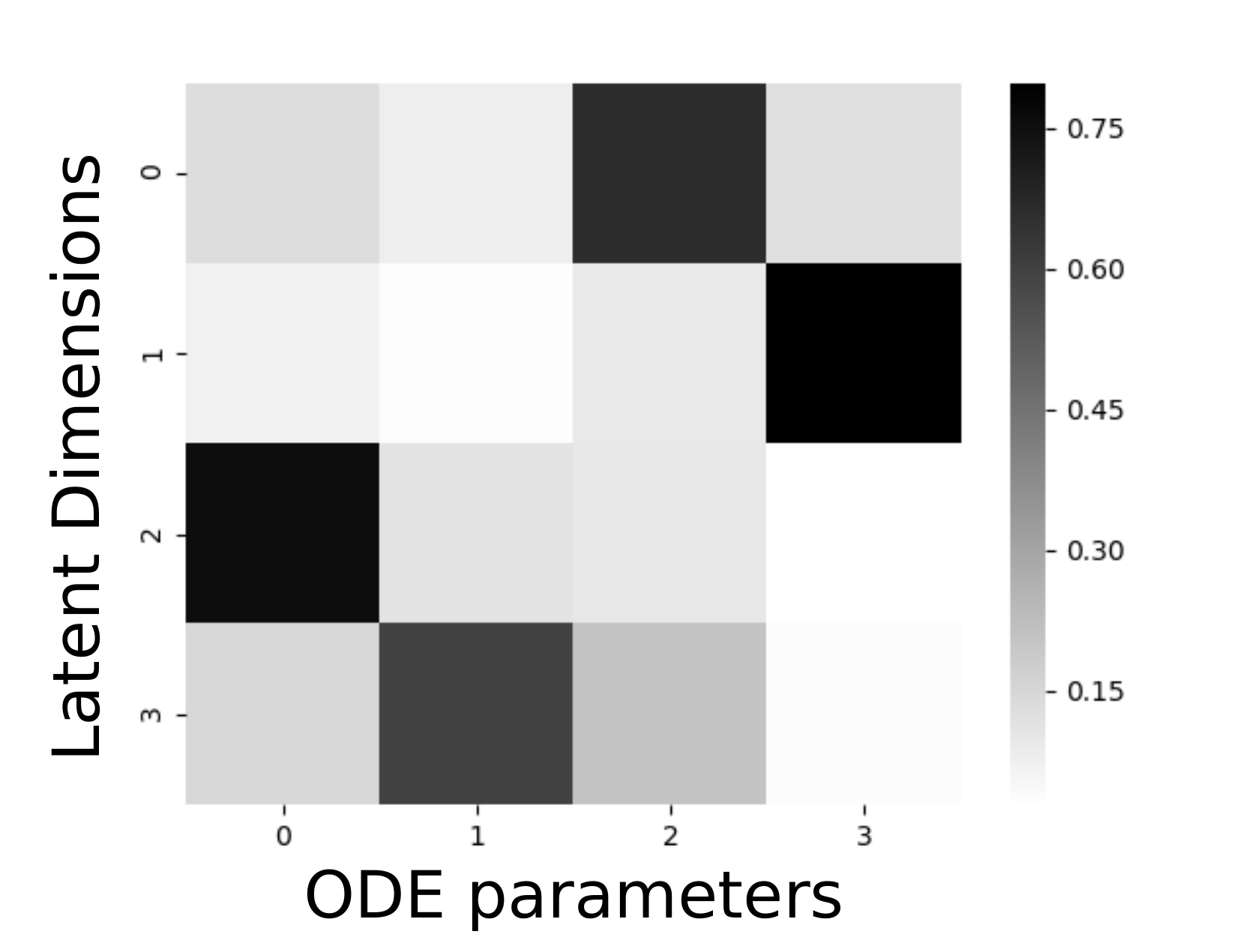}
      \caption{DSSM ($\delta=1$)}
    \end{subfigure}\hfill
    \begin{subfigure}[t]{.24\textwidth}
      \includegraphics[width=\linewidth]{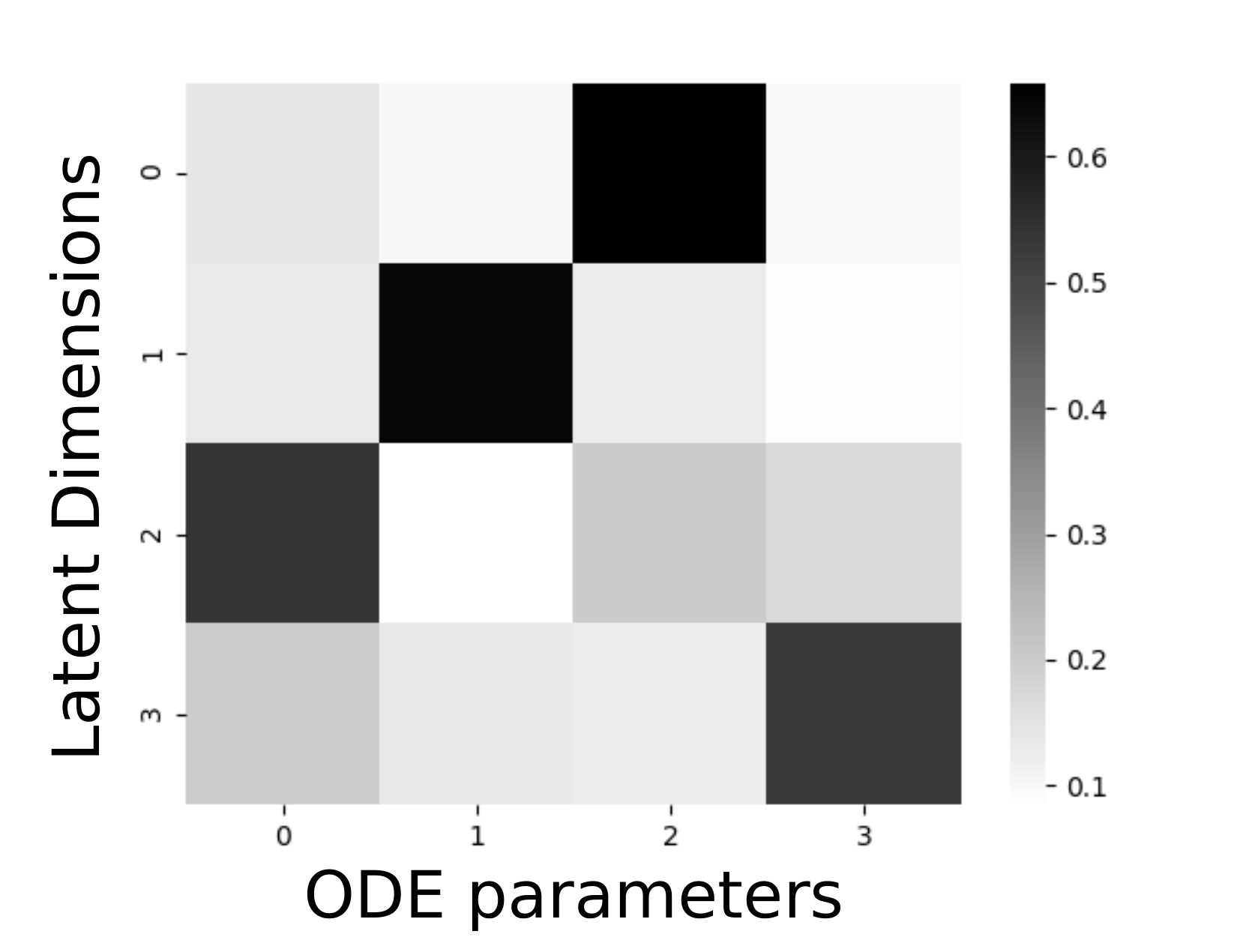}
      \caption{DSSM ($\delta=2$)}
    \end{subfigure}\hfill
    \caption{\textbf{Disentanglement in ODEs -- dependency matrices.} For different values of $\delta$, shown are the estimated dependencies between domain embeddings $D \in \mathcal{R}^4$ and ODE parameters $\alpha_{1-4}$ across all training sequences.}
    \label{fig:dm}
\end{figure}

\begin{wraptable}{r}{7cm}
\vspace{-1em}
\begin{tiny}
    \caption{Parameter inference for 100 realisations of noise with standard deviation of 0.1.}\label{tab-par-inf}
    \begin{tabular}{lcccc}
    \toprule
        & $\alpha_1$ & $\alpha_2$ & $\alpha_3$ & $\alpha_4$ \\
    \midrule
    Truth  & 2 & 1 & 4 & 1 \\
    DSSM ($\delta$=2)  & 2.01 $\pm$  0.04 & 0.97 $\pm$ 0.02 & 3.75 $\pm$ 0.05 & 1.0 $\pm$ 0.02 \\
    \bottomrule
    \end{tabular}
\end{tiny}
\end{wraptable} 
\paragraph{Disentanglement.}
Following~\cite{eastwood2018framework}, we trained a regressor to predict individual ODE parameters $\alpha_{1-4}$ from the 4-dimensional domain embeddings across training sequences.
Specifically, we used random forest and its ability to select important input features (as described in \cite{eastwood2018framework}), effectively evaluating disentanglement of the true generative factors of variations (ODE parameters) with respect to 4 latent dimensions.
The results given in Figure~\ref{fig:dm} show the effect of $\delta$ to the disentanglement.
For low values of $\delta$, generative factors tend to remain "entangled".
We then used trained regressor to perform explicit parameter inference on our benchmark trajectory.
Table~\ref{tab-par-inf} indicates that our method was able to almost fully recover the true ODE parameters.
Note that, even though the inference of $D$ is \emph{per se} static, we could in fact perform its estimation after each time step (can be thought of as Bayesian update). 
This may enable online parameter inference and could be crucial for some real-life applications e.g. in neuroscience~\citep{friston2003dynamic}.

\begin{table}[htbp]
    \caption{Mean squared error for 100 realisations of noise with standard deviation of 1.}
    \label{tab-mse}
    \begin{center}
    \begin{small}
    \begin{sc}
    \begin{tabular}{lccrr}
    \toprule
    Method & Prediction & Time & True functional form\\
    \midrule
    LSTM    & 24.7$\pm$ 30& online & not required\\
    GP      & 10.78$\pm$ 0 & 3 mins & not required\\
    SSM     & 3.56$\pm$ 5& online & not required\\
    DSSM ($\delta=0$) & 1.00$\pm$ 0.55& online & not required\\
    DSSM ($\delta=0.5$) & 0.81$\pm$ 0.43& online & not required\\
    DSSM ($\delta=1$) & 0.62$\pm$ 0.47& online & not required\\
    DSSM ($\delta=2$) & 0.38$\pm$ 0.42& online & not required\\
    FGPGM   & 0.15$\pm$ 0.5& 11 hours & required \\ 
    \bottomrule
    \end{tabular}
    \end{sc}
    \end{small}
    \end{center}
    \vspace{-1em}
\end{table}

\section{Experiment -- Learning Video Sequence Dynamics}
\label{sec:video}

We further test our framework on 2D bouncing ball videos where varying gravity values are applied across sequences.
We investigate the capability of our method to predict future frames of videos which contain previously unseen gravity values. We then explore the disentanglement of domain-specific factor (gravity direction) from generic dynamics (kinematic rules).
Finally, we show how fully generative DSSM can be used for uncontrolled and controlled video sequence generation~\footnote{More visualizations are found at: \url{https://sites.google.com/view/dssm}}.

\paragraph{Data set.}

Using the physics engine from \cite{fraccaro2017disentangled}
we rendered 51'200 = 16 $\times$ 3'200 video sequences in total,
where 16 stands for the number of gravity directions uniformly spread out across 2D space.
All gravity magnitudes were equal, and the initial ball states (including initial position and velocity) were randomly sampled.
We randomly chose two gravity directions and then singled out corresponding video sequences for validation (early stopping) and testing.
Our videos contained 70 binary images (30 used for domain recognition + 40 for video prediction) of dimensions 32 $\times$ 32. 

\begin{figure}[!t]
    \vspace{-2em}
    \centering
        \begin{subfigure}[t]{.49\textwidth}
	      \includegraphics[width=\linewidth]{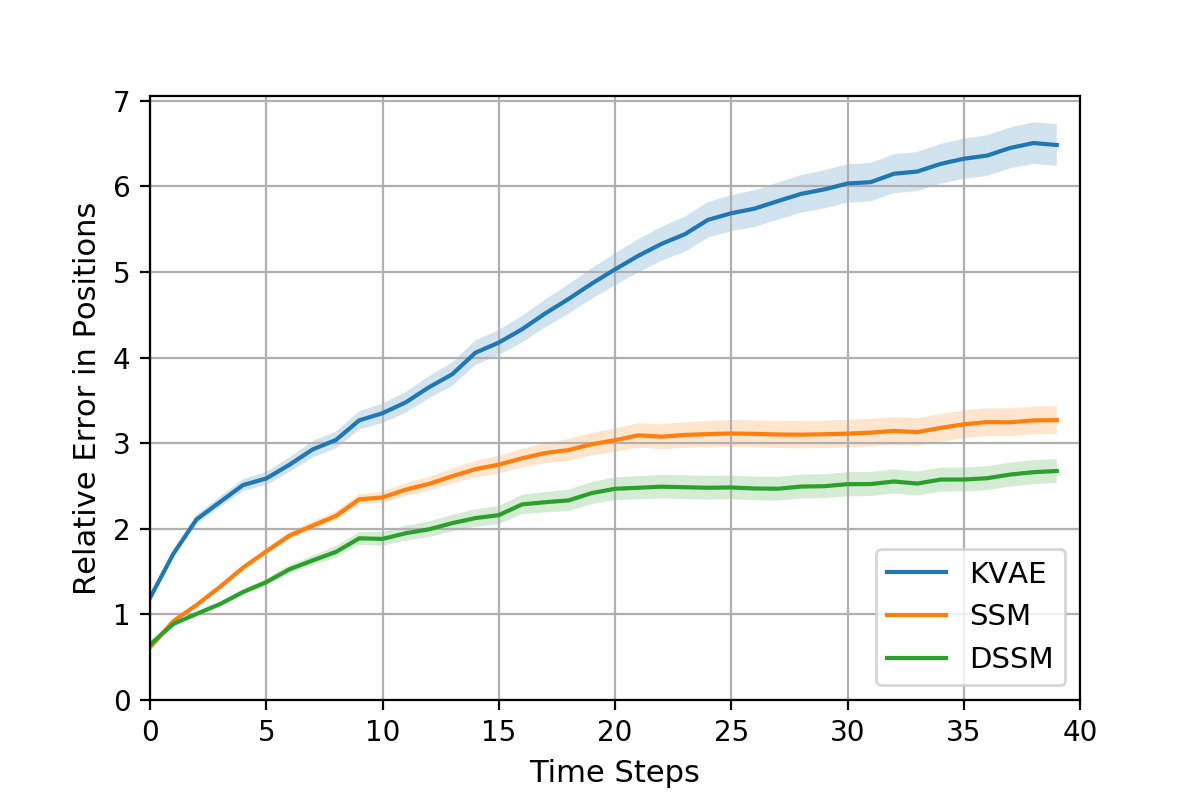}
        \end{subfigure}\hfill
        \begin{subfigure}[t]{.49\linewidth}
            \centering
            \includegraphics[width=.6\linewidth]{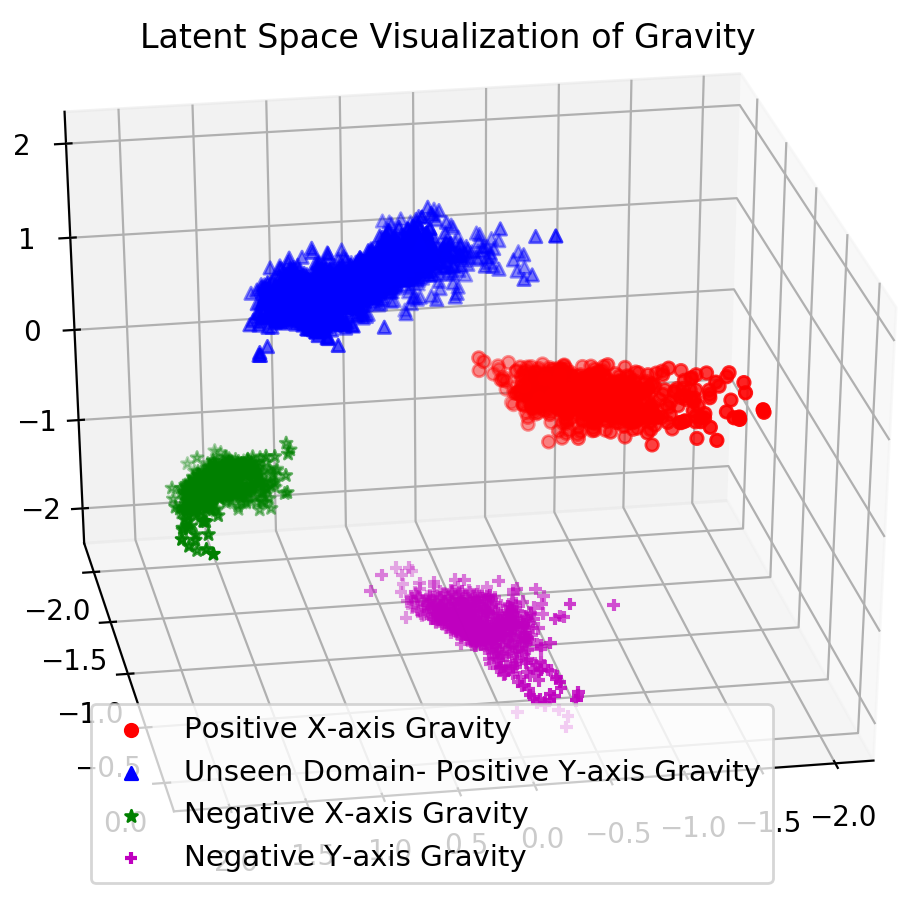}
        \end{subfigure}
        \caption{\emph{(left)} \textbf{Bouncing ball trajectory forecasting.} Comparison with respect to predicting future ball positions. Shown error curves are the test set averages. \emph{(right)} \textbf{Gravity embeddings in learned $D \in \mathcal{R}^3$ domain space.} Blue cluster originates from test sequences with previously unseen gravity.}
        \label{fig:prediction_embeddings}
        \vspace{-1em}
\end{figure}

\paragraph{Prediction.}
We performed long-term forecasting analysis (Figure~\ref{fig:prediction_embeddings}) comparing our method against state-of-the-art K-VAE \citep{fraccaro2017disentangled}, and also against a domain-free SSM ($D$ is fixed to 0) as an ablation study.
To detect ground truth ball position $p_t$, we used OpenCV~\footnote{https://opencv.org/} inbuilt functions (details in Appendix \ref{append-bball-detection})
We found that both of our models (DSSM and SSM) consistently outperformed K-VAE in predictions.
DSSM performed slightly better than SSM.
We suspect that because bouncing ball video dynamics is not as complex as ODE dynamics, the difference between the two was not as significant.

\paragraph{Disentanglement.}
To assess the disentanglement of the latent space, we embedded all training and test sequences into the 3-dimensional domain space, for visualization purposes.
The results are shown in Figure~\ref{fig:prediction_embeddings} for the 4 main gravity directions.
We observe very well-defined, compact clusters which are distributed in $\mathcal{R}^3$ in a meaningful way, resembling the topology of the 2D gravity space from our data generation procedure.
Recall also that the test sequences contain gravity values which were not seen during training time.
Nevertheless, DSSM remarkably managed to correctly place this new domain into the latent domain space, preserving the topology.

\begin{figure*}[h]
	\centering	
	\includegraphics[width=\textwidth]{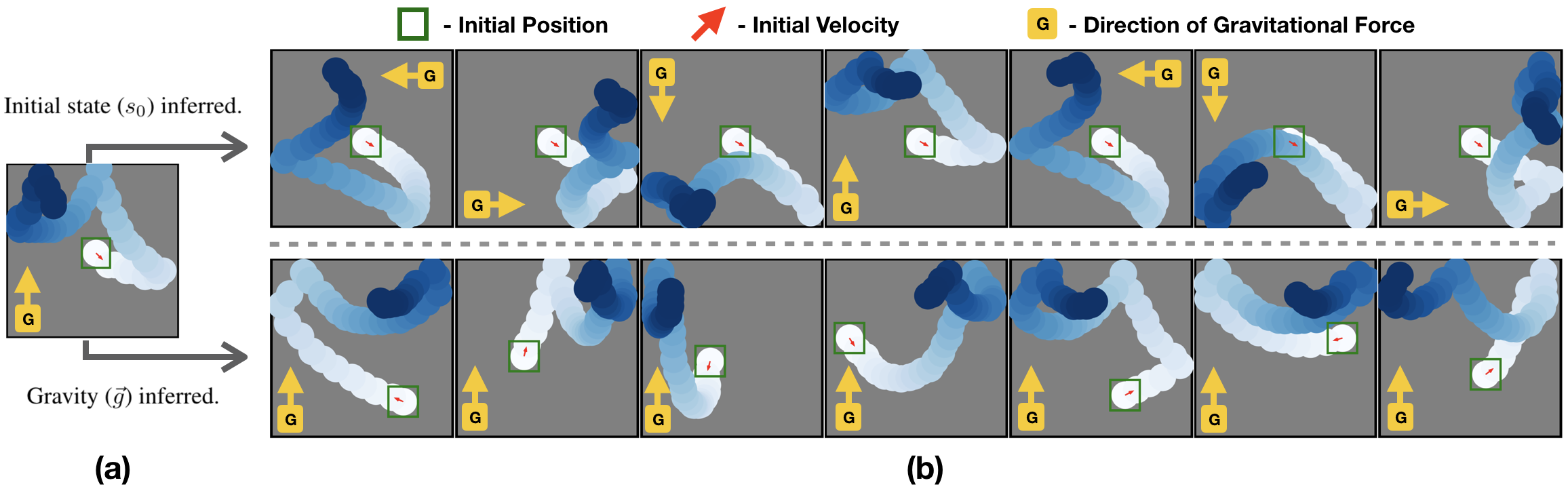}
	\caption{\textbf{Domain swapping and controlled generation.}
	\textbf{(a)} base sequence. \textbf{(b)} test sequences in which we injected (i) the gravity embedding; and (ii) the initial state of the base sequence.
	}
	\label{fig-controlled-generation}
\end{figure*}

\paragraph{Controlled and uncontrolled video generation.}
We performed interventions by "swapping gravity domains" between video sequences.
In other words, the latent representation of gravity was inferred from a base video sequence (using domain recognition network), and was then "injected" into a series of test sequences.
Similarly, we swapped the initial states.
The results are shown in Figure~\ref{fig-controlled-generation}.
Note that for the simplicity of visualizations, we focused only on four main gravity directions i.e. (+X,+Y),(-X,+Y),(+X,-Y) and (-X,-Y).
Finally, we also preformed uncontrolled video generation (see Figure~\ref{fig:uncontrolled_generation} in Appendix~\ref{a:video_other_plots}) simply by sampling from the priors $p_0(D)$ and $p_0(S_0)$.

\section{Conclusion}

In this work, we formalized a problem of learning sequence dynamics across heterogeneous domains, and then offered a concrete model (DSSM) and the corresponding VAE-based learning procedure.
On two separate domains, ODE and video sequences, we demonstrated the versatility of our approach. We performed cross-domain prediction, disentanglement, sequence generation and identification.
We also showed that our method is able to learn meaningful, compact latent spaces of sequences which might be crucial for applications which go beyond the scope of this paper e.g. learning cross-patient models in medical data or cross-environment dynamics models in robotics and reinforcement learning.
Furthermore, our work also sheds the light on the disentanglement and transfer learning paradigms from a novel perspective.
In the future, we plan to extend our work to include stochastic sequential systems, and also explore the applicability of our approach in the context of reinforcement learning.



\bibliographystyle{plain}
\bibliography{neurips_2019}
\newpage

\appendix

\section{Lower bound derivation (section~\ref{sec-bayes-filter})}
\label{sec-lowerbound}

\paragraph{Conditional log-likelihood term in $\mathcal{L}$}

\begin{align*}
            & \mathbb{D}_{q(\vec{S}|\vec{X})}[\text{log } p(\vec{X}|\vec{S})] = \mathbb{D}_{q(\vec{S}|\vec{X})}[\text{log } \prod_{i=1}^{T} p(X_i|S_i)] = \mathbb{D}_{q(\vec{S}|\vec{X})}[\sum_{i=1}^{T} \text{log } p(X_i|S_i)] = \sum_{i=1}^T \mathbb{D}_{q(S_i|\vec{X})}[\text{log }p(X_i|S_i)]
\end{align*}
\emph{(where the conditional independence follows from the state space model formulation)}

\paragraph{KL term in $\mathcal{L}$}

\small
\begin{align*}
            & \text{KL}(q(\vec{S}|\vec{X})||p_0(\vec{S})) \\
            &= \int_{\vec{S}} q(\vec{S}|\vec{X}) \text{ log } \frac{p_0(\vec{S})}{q(\vec{S}|\vec{X})} \\
            &= \int_{D} \int_{\beta} \int_{\vec{s}} q(S_0|\vec{X}) q(D|\vec{X}) q(\vec{\beta}|\vec{X},S_0,D)q(\vec{S}_0|S_0,D,\vec{\beta}) \\  
            &\text{ log } \frac{p_0(S_0)p_0(D)p_0(\vec{\beta}|S_0,D)p_0(\vec{S}_0|S_0,D,\vec{\beta})}{q(S_0|\vec{X})q(D|\vec{X})q(\vec{\beta}|\vec{X},S_0,D)q(\vec{S}_0|S_0,D,\vec{\beta})} \\
            & \text{\emph{(where we used the factorization of the variational and the prior distribution. $\vec{S}_0$ is vector $\vec{S}$ without $S_0$)}} \\
            &= \int_{D} q(D|\vec{X}) \text{ log } \frac{p_0(D)}{q(D|\vec{X})} \\
            &+ \int_{S_0} q(S_0|\vec{X}) \text{ log } \frac{p_0(S_0)}{q(S_0|\vec{X})} \\
            &+ \int_{D} \int_{\vec{\beta}} \int_{S_0} q(\vec{\beta}|\vec{X},D, S_0) \text{ log } \frac{p_0(\vec{\beta}|D, S_0)}{q(\vec{\beta}|\vec{X},D, S_0)}\\
            &+ \int_{D} \int_{\beta} \int_{\vec{S}} q(S_0|\vec{x}) q(D|\vec{x}) q(\vec{\beta}|\vec{X},S_0,D)q(\vec{S}_0|S_0,D,\vec{\beta}) \text{ log } \frac{p_0(\vec{S}_0|S_0,D,\vec{\beta})}{q(\vec{S}_0|S_0,D,\vec{\beta})}\\
            & \text{\emph{(where we dropped the integral sums for which the corresponding term does not depend on)}} \\
            &= \text{KL}(q(D|\vec{X})||p_0(D)) \\
            &+ \text{KL}(q(S_0|\vec{X})||p_0(S_0)) \\
            &+ \text{KL}(q(\vec{\beta}|\vec{X},D, S_0)||p_0(\vec{\beta}|D, S_0)) \\
            & \text{\emph{(where the last term vanishes since $\vec{s}_0|s_0,D,\vec{\beta}$ is deterministic)}} \\
            &= \text{KL}(q(D|\vec{X})||p_0(D)) \\
            &+ \text{KL}(q(S_0|\vec{X})||p_0(S_0)) \\
            &+ \sum_{i=1}^T \underset{q(\beta_i,D,S_{i-1}|\vec{x})}{\mathbb{D}}\text{KL}(q(\beta_i|X_i,D, S_{i-1})||p_0(\beta)) \\
            & \text{\emph{(where we have $p_0(\beta_i|D, s_i)=p_0(\beta)$ by design)}} \\
\end{align*}
\normalsize

\newpage

\section{Learning algorithm (section~\ref{sec-bayes-filter})}
\label{a:la}

\begin{algorithm}[htbp]
   \caption{One iteration of the training procedure}
   \label{alg-training}
\begin{algorithmic}
   \STATE {\bfseries Input:} sequence $\vec{X}$ of length $T$
   \STATE $[\mu^D, \Sigma^D] = \phi_D^{enc}(\vec{X}), \quad D \sim \mathcal{N}(\mu^D, \Sigma^D)$
   \STATE $[\mu^S, \Sigma^S] = \phi_S^{enc}(\vec{X}), \quad S_0 \sim \mathcal{N}(\mu^S, \Sigma^S)$
   \FOR{$i=1$ {\bfseries to} $T$}
        \STATE \emph{Predict step:} $S_i^- = f(S_{i-1}, D)$
        \STATE \emph{Estimate residual:} $[\mu^\beta_i, \Sigma^\beta_i] = \phi_\beta^{enc}(S_i^-, X_i), \quad \beta_i \sim \mathcal{N}(\mu^\beta_i, \Sigma^\beta_i)$
        \STATE \emph{Update step:} $S_i = S_i^- + \beta_i$
        \STATE \emph{Predict observation:} $\hat{X_i} = g(S_i)$
   \ENDFOR
   \STATE ll\_loss = LL$(\vec{X}, \vec{\hat{X}})$ (see Eq~(\ref{eq-ll}))
   \STATE kl\_loss = KL$(D, S_0, \vec{\beta})$; (see Eq~(\ref{eq-kl}))
   \STATE mm\_loss = MM$(\phi_D^{enc}(\vec{X}))$; (see Eq~(\ref{eq-mm}))
   \STATE \emph{Backpropagate(ll\_loss, kl\_loss, mm\_loss)}
\end{algorithmic}
\end{algorithm}

\section{Lotka-Volterra experiments (section~\ref{sec:dynamical_system})}
\label{a:lv}

\subsection{Details}

We discuss the remaining details of DSSM training and architecture used in the experiments on Lotka-Volterra ODE system. First of all, we set all the hidden states to be of equal size 80. This includes the actual state $S$ and the hidden states of used multi layer perceptrons (with ReLU activation units) and LSTMs.
The size of $D$ was set to 4. All LSTMs had a depth of two layers. For optimization, we used Adam with the learning rate of 0.001, and a learning rate decay of 0.94 which was applied after each epoch. We used batches of size 50 and trained our model for 200 epochs, but due to early stopping applied on held-out-set of data, it converged much earlier (50 epochs were roughly enough). We also added white Gaussian noise with standard deviation of $0.5$ to all training samples, in order to improve the robustness of the model in the test time. Finally, we followed the simple linear KL annealing scheme. Namely, starting from weight 0, we increased the weight of the corresponding term by 0.000001 in each iteration until it reached 1, and formed the correct lower bound from Eq~(\ref{eq-lb}). The remaining relevant details are given in the main text and can in addition be found in our code repository which will be made publicly available upon acceptance. For the baselines we used the official repository of FGPGM~\footnote{https://github.com/wenkph/FGPGM} and the modification of it to fit a GP. For LSTMs, we used 2-layered architecture with hidden states of size 80.

\section{Bouncing ball experiments (section~\ref{sec:video})}

\subsection{Details}

Most of the training details were similar to our ODE experiment, so we will summarize the differences only.
To get compressed representation of each frame, the images were first passed through a shallow convolutional network. Kernel size was set to 3x3, while the network depth was 64. The step size was 1 in both directions. 
All of the hidden latent states were equal to 64. 
To parameterize $g$ we used a deconvolutional network with transposed convolutions. 
The kernel size was set to 5.
Similarly to \citep{fraccaro2017disentangled}, we also found that down-weighing the reconstruction term helps in faster convergence. In particular we downscaled the conditional log-likelihood from Eq (\ref{eq-ll}) by a factor of 10.
We use ADAM as the optimizer with 0.0008 as the initial learning rate, and weight decay of 0.6 applied every 20 epochs.


 \lstset{frame=tb,
  language=Python,
  aboveskip=3mm,
  belowskip=3mm,
  showstringspaces=false,
  columns=flexible,
  basicstyle={\small\ttfamily},
  numbers=none,
  numberstyle=\tiny\color{gray},
  keywordstyle=\color{blue},
  commentstyle=\color{dkgreen},
  stringstyle=\color{mauve},
  breaklines=true,
  breakatwhitespace=true,
  tabsize=3
}
 
 \subsection{Algorithm for Detecting Ball Positions}
 \label{append-bball-detection}
 We used OpenCV inbuilt functions to detect the pixel level positions of the ball in the images. The algorithm follows.
\begin{lstlisting}
import cv2
import imutils
def find_positions(image):
    ret, binary_mask = cv2.threshold(image, 0.01, 1, cv2.THRESH_BINARY)
    binary_mask = cv2.erode(binary_mask, None, iterations=1)
    binary_mask = cv2.dilate(binary_mask, None, iterations=1)
    fake_frame = cv2.convertScaleAbs(binary_mask.copy())
    cnts = cv2.findContours(fake_frame,
                        cv2.RETR_EXTERNAL,
                        cv2.CHAIN_APPROX_SIMPLE)
    cnts = imutils.grab_contours(cnts)
    c = max(cnts, key=cv2.contourArea)
    ((x, y), radius) = cv2.minEnclosingCircle(c)
    return x, y
\end{lstlisting}

\subsection{Other plots}
\label{a:video_other_plots}

\begin{figure}[htbp]
	\centering	
	\includegraphics[width=\textwidth]{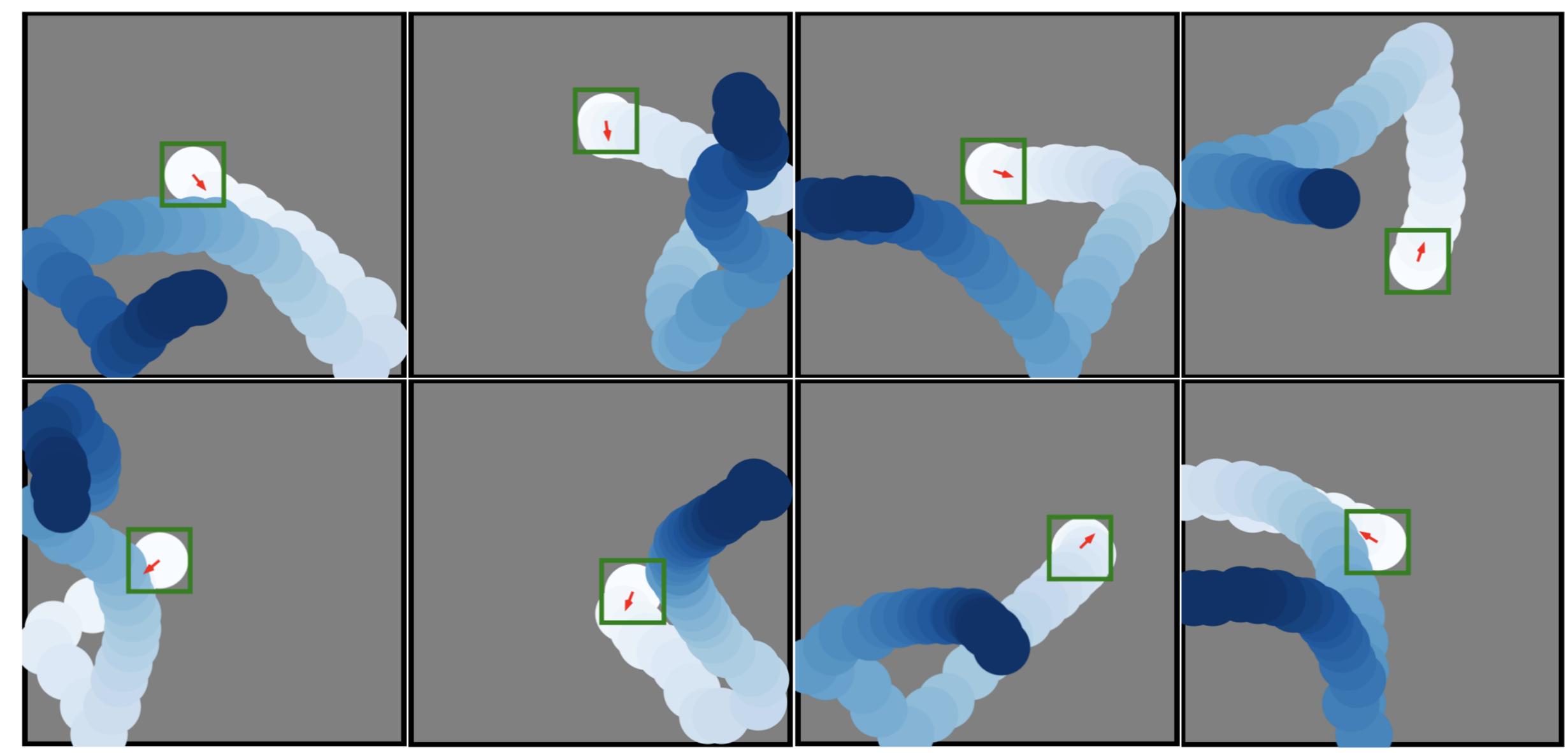}
	\caption{\textbf{uncontrolled video generation.} Both gravity and initial states are sampled from prior.}
	\label{fig:uncontrolled_generation}
\end{figure}

\end{document}